\documentclass[runningheads]{llncs}
\usepackage[T1]{fontenc}
\usepackage{graphicx}
\usepackage{amsmath}
\usepackage{amssymb}
\usepackage{booktabs}
\usepackage{multirow}
\usepackage{xcolor}
\usepackage{marvosym}
\usepackage{algorithm}
\usepackage{algpseudocode}
\usepackage{color}
\usepackage{xspace}
\usepackage{tikz}
\usepackage{pgfplots}
\pgfplotsset{compat=1.18}
\usetikzlibrary{positioning, shapes.geometric, arrows.meta, calc, fit, backgrounds}
\usepackage[colorlinks=true,
            linkcolor=blue!60!black,
            citecolor=blue!60!black,
            urlcolor=blue!60!black,
            filecolor=blue!60!black,
            menucolor=blue!60!black,
            allbordercolors={1 1 1}]{hyperref}

\newcommand{\method}{\textsc{Savor}\xspace}
\definecolor{halgreen}{RGB}{34,139,34}
\definecolor{halred}{RGB}{200,40,40}
\begin{document}
\title{SAVOR: Self-Aware Visual Grounding via\\Confidence-Calibrated Reinforcement Learning\\for Multimodal Hallucination Mitigation}
\titlerunning{SAVOR: Self-Aware Visual Grounding via Calibrated RL}
\author{Zixiu Ding\inst{1,2} \and Zilin Zhao\inst{2} \and Yingjie He\inst{2} \and Xinlang Kang\inst{2} \and Guansu Wang\inst{2} \and Wei Zhang\Letter}
\authorrunning{Z. Ding et al.}
\institute{Central South University \and Peking University}
\maketitle
\begin{abstract}
Multimodal large language models (MLLMs) have made strong progress on
visual question answering and image captioning, yet they still produce
fluent claims about objects, attributes, or relations that are not grounded
in the image. Many remedies either modify decoding at test time, which
adds latency, or fine tune with preferences such as DPO variants, which
teach which answer is preferred but not when the model's own answer is
unreliable. We argue that calibrated self assessment is the missing signal.
We introduce \method, a training framework that (i) augments the output
schema with token and answer confidence, (ii) optimises the policy with a
Group Relative Policy Optimisation (GRPO) objective that penalises
calibration error and poor abstention decisions, and (iii) uses the
learned confidence at inference time to revisit visual evidence only when
the model is uncertain. Experiments on POPE, HallusionBench, AMBER and
MMHal-Bench across two recent backbones (InternVL3-8B and Qwen3-VL-8B)
show that \method reduces hallucination while preserving general capability
on MME and MMBench, with lower Expected Calibration Error than DPO and
decoding baselines.
\end{abstract}
%
\section{Introduction}
\label{sec:intro}

Multimodal large language models (MLLMs) such as
InternVL3~\cite{zhu2025internvl3}, Qwen3-VL~\cite{bai2025qwen3} and
LLaVA-OneVision-1.5~\cite{an2025llava} have narrowed the gap between
vision and language, and now perform well on many visual question
answering and captioning benchmarks~\cite{kang2026hssbench,luo2024codis}.
A persistent obstacle, however, is
that these models still produce fluent and confident claims about objects,
attributes, counts, or relations that are not supported by the image
\cite{li2023evaluating,guan2024hallusionbench,wang2023amber,kang2026modality}. Such errors are especially
problematic because hallucinated answers often carry the same surface
certainty as correct ones, leaving users with little signal for when an
answer should be trusted~\cite{yu2026believing,lou2026helpers}. In
practice, the model may fail twice: it gives
an unsupported visual statement, then presents it with enough confidence
that a user or downstream system has no clear reason to reject it.

Existing mitigation methods mostly follow one of two routes. Test-time
methods, including contrastive decoding~\cite{leng2024mitigating}, attention
reallocation~\cite{huang2024opera}, and verifier-style correction~\cite{chen2024halc}, intervene during
generation without changing the model parameters. They can reduce
hallucination, but they add latency and do not change the policy that
produced the error. Preference tuning methods such as RLHF-V~\cite{yu2024rlhf},
HA-DPO~\cite{zhao2023beyond}, HALVA~\cite{sarkar2024mitigating}, and mDPO~\cite{wang2024mdpo}
instead train on hallucination-oriented preference pairs. These methods
internalise part of the correction, but their supervision is still
comparative: the model learns which answer is preferred, not how reliable
its own answer is. As a result, a policy can become more accurate on
average while remaining poorly calibrated. This distinction matters because
a model that hallucinates less can still be risky if it is most confident
exactly when it is wrong.

This paper studies hallucination mitigation through calibrated self
assessment. The idea is motivated by work on textual LLMs showing that
models can express meaningful uncertainty and that explicit supervision of
uncertainty can improve truthfulness
\cite{lin2022teaching,kadavath2022language,tian2023just,xu2024sayself}. The multimodal setting
makes this direction natural. Visual hallucinations can often be checked
against the input image, giving a relatively clean correctness signal, and
current MLLMs already support structured reasoning traces in which a
confidence channel can be inserted without architectural changes. The key
question is how to make the accuracy of the model's own self assessment
part of the optimisation target, instead of treating confidence as a
diagnostic added after generation. In other words, the model should not
only learn to give a better answer; it should also learn when its answer is
visually supported, when it is uncertain, and when abstention is safer than
a guess.

We introduce \method, a training and inference framework for making an
MLLM both more grounded and better calibrated. During a short supervised
initialisation stage, the model learns to emit confidence values for
visually grounded spans, together with a single confidence score for the
whole answer. The policy is then optimised with Group Relative Policy
Optimisation (GRPO)~\cite{shao2024deepseekmath}. In addition to answer correctness,
the reward includes a calibration term that compares confidence with
empirical rollout accuracy, and an abstention term that discourages
confident guesses when the visual evidence is weak. At inference time, the
same answer confidence triggers a lightweight visual re-attention step:
when confidence is below a threshold, the model crops the most attended
region, queries itself again, and either returns the revised answer or
abstains. This keeps the correction loop inside the same policy, instead
of relying on a separate verifier or applying an expensive decoding
strategy to every prompt.

Our experiments are designed to test both hallucination reduction and the
quality of the learned confidence signal. Across POPE, HallusionBench,
AMBER and MMHal-Bench, \method consistently improves over
preference-tuning and standard GRPO baselines on two recent 8B-scale
backbones. The gains are most visible in calibration: on the adversarial
split of POPE, for instance, Expected Calibration Error is reduced from
$0.272$ with vanilla GRPO to $0.085$ with \method. At the same time,
general visual-language ability, as measured by MME and MMBench, remains
essentially unchanged. These results suggest that the calibration objective
improves reliability without turning the method into a narrow trade-off
against general capability.

The main contributions are summarised as follows:
\begin{itemize}
\item We formulate MLLM hallucination mitigation as a calibration problem,
rather than only a preference ranking problem, and use calibration error as
a direct reward signal in multimodal RL.
\item We develop \method, a unified pipeline that learns to answer, report
visual confidence, abstain under weak evidence, and revisit visual evidence
within the same policy.
\item We evaluate the approach on InternVL3-8B and Qwen3-VL-8B across
hallucination, calibration and general capability benchmarks, showing
consistent hallucination reduction and substantially lower ECE while
preserving overall capability.
\end{itemize}

\section{Related Work}
\label{sec:related}

\subsection{Hallucination Diagnosis and Correction at Inference}
Object, attribute and relational hallucinations in MLLMs are commonly
measured by POPE~\cite{li2023evaluating}, HallusionBench~\cite{guan2024hallusionbench},
AMBER~\cite{wang2023amber} and MMHal-Bench~\cite{sun2024aligning}. These benchmarks cover
object existence, counting, attributes, spatial relations and visually
grounded commonsense; related suites probe cross-modal
ambiguity~\cite{wang2025mucar}, context dependence~\cite{luo2024codis},
cross-view spatial reasoning~\cite{feng2026views}, temporal
grounding~\cite{zhu2026comet} and robustness to structural
corruption~\cite{kang2026modality}. Diagnostic studies further connect such errors to
language priors and attention concentration~\cite{huang2024opera,liu2024reducing}. This line of work
establishes our evaluation setting, but mostly measures hallucination after
generation; it does not teach the model to recognise when its own visual
claims are unreliable.

Correction methods used at inference can intervene without updating the
model: VCD~\cite{leng2024mitigating} uses visual contrastive logits,
OPERA~\cite{huang2024opera} adjusts attention, and HALC~\cite{chen2024halc} adds verification.
Although these methods can improve frozen MLLMs, the correction remains
external to the policy and adds decoding or verification cost at deployment
time. \method keeps a lightweight correction loop, but triggers it only
when the learned confidence signal indicates uncertainty.

\subsection{Preference and Reinforcement Learning for Multimodal Alignment}
Hallucination-aware alignment methods update the model with preference or
RL signals. RLHF-V~\cite{yu2024rlhf}, HA-DPO~\cite{zhao2023beyond}, HALVA~\cite{sarkar2024mitigating}
and mDPO~\cite{wang2024mdpo} use human or synthetic preference data to make
visual instruction tuning less prone to hallucination. Their supervision,
however, is comparative: the model learns that one answer is better than
another, not that its reported confidence should match the probability of
being correct. Thus preference tuned policies may become more accurate
while remaining overconfident on ambiguous prompts. A parallel line
reshapes the reward itself, aligning updates with gradient
evidence~\cite{zheng2026gradients}, spreading group relative credit over
intermediate steps~\cite{shi2026spader}, or replacing surface overlap
scores with hierarchical task aware rewards for grounded
generation~\cite{wang2026ngrams}.

GRPO, introduced in DeepSeekMath~\cite{shao2024deepseekmath} and scaled in
DeepSeek-R1, removes the critic by normalising rewards
within a rollout group. Recent multimodal variants apply similar ideas to
visual reasoning~\cite{wang2026vl} and hallucination suppression
\cite{wang2025mitigating}. We also use GRPO, but the rollout group serves an
additional role: it supplies an empirical accuracy estimate against which
the model's verbalised confidence can be calibrated. Because the group
mixes agreeing and conflicting rollouts, weighting signals by their mutual
consistency rather than summing them is the same principle used to reconcile
conflicting updates in distributed training~\cite{hong2026conflict}.

\subsection{Calibration, Verbalised Uncertainty and Self-aware Policies}
Textual LLM studies show that models can verbalise useful uncertainty
\cite{lin2022teaching}, encode known/unknown signals~\cite{kadavath2022language}, and be
improved by prompting or supervised calibration~\cite{tian2023just,xu2024sayself}.
Related diagnostics read reliability off internal dynamics rather than
stated confidence, for instance through entropy
trajectories~\cite{zhu2026edis}, the spread of competing
solutions~\cite{jiang2026foe}, sensitivity to input
order~\cite{kang2026orderprobe}, or explicit reasoning
pathways~\cite{dong2026neureasoner}.
Work on self-rewarding models further suggests that model judgements can
enter the optimisation loop~\cite{yuan2024self}. These results motivate a
readable confidence channel that requires no architectural change and can
directly control inference decisions.

For MLLMs, calibration must be tied to visual grounding: a model may be
confident because of language priors even when the image contradicts them.
Existing calibration work rarely targets this visual failure mode, while
hallucination mitigation rarely optimises calibration as a first-class
objective. \method connects the two by rewarding visually grounded
confidence and using that confidence to decide when self-correction should
be attempted.

\section{Method}
\label{sec:method}

\subsection{Overview}
\label{sec:method:overview}

\begin{figure}[t]
\centering
\includegraphics[width=0.86\textwidth]{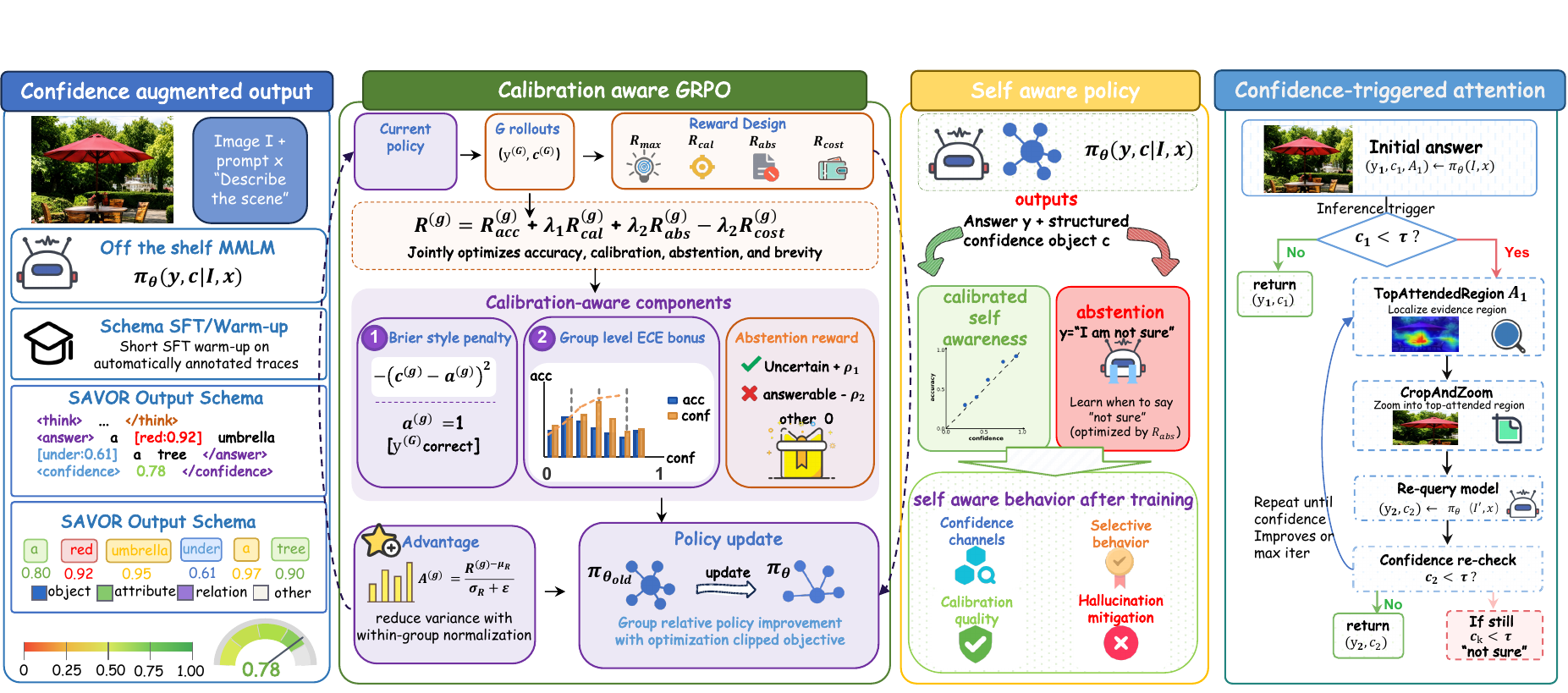}
\caption{Overview of \method. The model learns confidence outputs, is optimised with a calibration reward in GRPO, and uses low confidence to run one crop-and-zoom query before answering or abstaining.}
\label{fig:overview}
\end{figure}

\method starts from an MLLM $\pi_\theta(y\mid I,x)$, where $I$ is an image
and $x$ a textual prompt, and turns it into a policy
$\pi_\theta(y, c \mid I, x)$ that emits both an answer $y$ and a structured
confidence object $c$. Training proceeds in two stages (Fig.~\ref{fig:overview}):
a brief supervised stage that teaches the output schema, followed by GRPO
with a calibration reward that optimises both the answer and the confidence
under image grounded rewards. At inference time, a low confidence score
gates an optional visual re-attention step (\S\ref{sec:method:reatt}).

\subsection{Output Schema with Visual Confidence}
\label{sec:method:schema}

For an image--prompt pair $(I,x)$, the original MLLM defines a policy
$\pi_\theta(y\mid I,x)$ over textual answers. We augment the output space
so that the policy produces not only an answer sequence
$y=(w_1,\ldots,w_T)$, but also a structured confidence object
\begin{equation}
 c = \bigl(c_{\mathrm{ans}},\{(s_m,c_m)\}_{m=1}^{M}\bigr),
 \qquad c_{\mathrm{ans}},c_m\in[0,1],
 \label{eq:conf-object}
\end{equation}
where $c_{\mathrm{ans}}$ is the confidence of the whole answer and each
$(s_m,c_m)$ pairs a visually grounded span $s_m\subseteq y$ with its visual
confidence. The spans cover objects, attributes, counts and spatial
relations that should be checkable from the image. The resulting policy is
therefore
\begin{equation}
 \pi_\theta(o\mid I,x)
 = \pi_\theta(y,c_{\mathrm{ans}},\{(s_m,c_m)\}_{m=1}^{M}\mid I,x),
 \qquad o=(y,c),
 \label{eq:aug-policy}
\end{equation}
which makes answer generation and confidence estimation part of the same
sequence decision, rather than two separate steps after decoding. The
concrete text serialisation is illustrated in Fig.~\ref{fig:overview}; we
do not require any additional confidence head or modification to the vision
encoder.

The first training stage teaches this augmented output space with
supervised fine tuning. Let
$\mathcal{D}_{\mathrm{sft}}=\{(I_i,x_i,y_i^\star,c_i^\star)\}_{i=1}^{N}$
be the initialisation set, where $c_i^\star$ contains pseudo labelled span
and answer confidences. Given $K$ stochastic responses from an ensemble or
from repeated sampling, a span pseudo label is estimated as
\begin{equation}
 c_{i,m}^{\star}=\frac{1}{K}\sum_{k=1}^{K}
 \mathbb{1}\bigl[s_{i,m}\text{ is supported in }y_i^{(k)}\bigr],
 \label{eq:span-pseudo}
\end{equation}
and the answer target is computed analogously from answer agreement or
task correctness when available. The SFT objective combines standard
response likelihood with confidence regression after parsing the verbalised
confidence values:
\begin{equation}
 \mathcal{L}_{\mathrm{sft}}
 = -\sum_{t=1}^{T_i}\log \pi_\theta(w_{i,t}^{\star}\mid I_i,x_i,w_{i,<t}^{\star})
 + \alpha\bigl(c_{\mathrm{ans},i}-c_{\mathrm{ans},i}^{\star}\bigr)^2
 + \gamma\frac{1}{M_i}\sum_{m=1}^{M_i}(c_{i,m}-c_{i,m}^{\star})^2 .
 \label{eq:sft}
\end{equation}
This stage does not aim to solve hallucination by itself; it only makes
the output format stable enough for the subsequent RL stage to optimise
confidence as a behavioural signal.

\subsection{GRPO with a Calibration Reward}
\label{sec:method:rl}

\paragraph{Policy and rollout.}
After schema initialisation, we optimise the augmented policy with GRPO and
a calibration reward. For each prompt $(I,x)$, the frozen reference policy
$\pi_{\mathrm{ref}}$ and the previous policy $\pi_{\theta_{\mathrm{old}}}$
are kept fixed during one update. We sample a group of $G$ rollouts
$\mathcal{G}=\{o^{(g)}\}_{g=1}^{G}$, where
$o^{(g)}=(y^{(g)},c^{(g)})\sim\pi_{\theta_{\mathrm{old}}}(\cdot\mid I,x)$.
Each rollout is parsed into an answer $y^{(g)}$, an answer confidence
$c_{\mathrm{ans}}^{(g)}$, span confidences $\{c_m^{(g)}\}$, and a task
correctness signal $a^{(g)}\in[0,1]$. The group mean and standard deviation
of rewards are used to form the relative advantage
\begin{equation}
 A^{(g)}=\frac{R^{(g)}-\mu_R}{\sigma_R+\epsilon},\qquad
 \mu_R=\frac{1}{G}\sum_{j=1}^{G}R^{(j)},\quad
 \sigma_R^2=\frac{1}{G}\sum_{j=1}^{G}(R^{(j)}-\mu_R)^2 .
 \label{eq:advantage}
\end{equation}
This relative normalisation is important in our setting because prompts
vary substantially in difficulty: a hard visual reasoning prompt should
not be compared directly with an easy object query, but its rollouts can
still be ranked against each other.

The policy update uses the GRPO clipped objective~\cite{schulman2017proximal} with a KL penalty to the
reference model:
\begin{equation}
\begin{aligned}
\mathcal{J}_{\mathrm{grpo}}(\theta)
=\frac{1}{G}\sum_{g=1}^{G}
\min\Bigl(&r_\theta^{(g)} A^{(g)},
\operatorname{clip}(r_\theta^{(g)},1-\epsilon_c,1+\epsilon_c) A^{(g)}\Bigr) \\
&-\eta_{\mathrm{kl}}\,
D_{\mathrm{KL}}\bigl(\pi_\theta(\cdot\mid I,x)\,\|\,\pi_{\mathrm{ref}}(\cdot\mid I,x)\bigr),
\end{aligned}
\label{eq:grpo}
\end{equation}
where
\begin{equation}
 r_\theta^{(g)}=
 \frac{\pi_\theta(o^{(g)}\mid I,x)}
 {\pi_{\theta_{\mathrm{old}}}(o^{(g)}\mid I,x)} .
 \label{eq:ratio}
\end{equation}
The KL term prevents the confidence objective from drifting into a narrow
policy tuned only for hallucination benchmarks, while the clipped ratio
stabilises updates when a rollout receives a large calibration or
abstention reward.

\paragraph{Reward design.}
The total reward for each rollout is
\begin{equation}
R^{(g)} \;=\; R_{\text{acc}}^{(g)}
\;+\;\lambda_1\, R_{\text{cal}}^{(g)}
\;+\;\lambda_2\, R_{\text{abs}}^{(g)}
\;+\;\lambda_3\, R_{\text{span}}^{(g)}
\;-\;\lambda_4\, R_{\text{cost}}^{(g)},
\label{eq:reward}
\end{equation}
with the five terms defined as follows.

\smallskip
\noindent\textbf{Accuracy reward.}
$R_{\text{acc}}^{(g)} = \mathbb{1}[\text{correct}(y^{(g)})]$ for
discriminative tasks such as POPE yes/no and VQA exact match, and a
normalised generative score for open answer tasks, where higher AMBER F1
and lower CHAIR both receive larger rewards.

\smallskip
\noindent\textbf{Calibration reward.}
Each rollout's answer confidence $c_{\mathrm{ans}}^{(g)}$ is matched
against the realised correctness $a^{(g)}\in\{0,1\}$ via a Brier penalty,
augmented by an Expected Calibration Error (ECE) bonus at the group level:
\begin{equation}
R_{\text{cal}}^{(g)} \;=\; -\,\bigl(c_{\mathrm{ans}}^{(g)} - a^{(g)}\bigr)^2
\;+\;\beta\,\bigl(\mathrm{ECE}_{\text{group,prev}} - \mathrm{ECE}_{\text{group}}\bigr),
\label{eq:calreward}
\end{equation}
where $\mathrm{ECE}_{\text{group}}$ is computed over the $G$ rollouts of
the current prompt by binning $c_{\mathrm{ans}}^{(g)}$ into $K$ buckets and
comparing accuracy and mean confidence in each bucket. The first term
shapes individual rollouts; the bracketed difference gives group credit to
updates that reduce calibration error compared with the previous group.

\smallskip
\noindent\textbf{Abstention reward.}
We also allow the policy to emit a designated abstention answer
$y_\varnothing=\text{``I am not sure.''}$. We define
\begin{equation}
R_{\text{abs}}^{(g)} \;=\;
\begin{cases}
+\rho_1, & \text{if }y^{(g)}=y_\varnothing\text{ and the prompt is
intrinsically uncertain},\\
-\rho_2, & \text{if }y^{(g)}=y_\varnothing\text{ but the prompt is
answerable},\\
0, & \text{otherwise},
\end{cases}
\end{equation}
where ``intrinsically uncertain'' means that most rollouts in the group are
incorrect and the group accuracy is below a threshold $\tau_u$. The
abstention reward is therefore supervised by the rollout group itself,
removing the need for explicit ``unanswerable'' labels.

\smallskip
\noindent\textbf{Span grounding reward.}
Confidence values attached to generated spans should also reflect whether
those spans are visually supported. Let $v_m^{(g)}\in\{0,1\}$ denote
whether span $s_m^{(g)}$ is verified by the available training annotations,
object tags, OCR evidence, or lightweight grounding heuristics. We define
\begin{equation}
R_{\text{span}}^{(g)}
= -\frac{1}{M_g}\sum_{m=1}^{M_g}\bigl(c_m^{(g)}-v_m^{(g)}\bigr)^2,
\label{eq:spanreward}
\end{equation}
so that unsupported visual spans are penalised when assigned high
confidence, while genuinely grounded spans are not discouraged. This term
connects answer calibration with the local visual grounding signal used by
the re-attention module.

\smallskip
\noindent\textbf{Length cost.}
$R_{\text{cost}}^{(g)} = |y^{(g)}|/L_{\max}$ discourages reward hacking
via verbosity.

\paragraph{Why calibration as a direct reward?}
Standard RLHF~\cite{ouyang2022training} / DPO~\cite{rafailov2023direct} objectives are rank based: they tell the model that answer
$A$ should beat answer $B$, but never that the model's own internal
scoring of $A$ should be numerically correct. As a result, policies after
training are often more accurate but also more overconfident
\cite{kadavath2022language}. The Brier term in Eq.~\eqref{eq:calreward} penalises
both directions of miscalibration symmetrically, while the ECE bonus
aggregates the signal at the group level and damps the variance of
individual Brier estimates. We show in \S\ref{sec:exp:ablation} that
removing either subterm visibly degrades calibration without changing
accuracy much, isolating the contribution of this design.

\subsection{Visual Re-Attention Triggered by Confidence}
\label{sec:method:reatt}

After RL training, $c_{\mathrm{ans}}$ is a usable scalar in $[0,1]$ at
inference time. We use it to decide whether one extra correction pass is
needed. The decision rule is
\begin{equation}
\hat{o}=\begin{cases}
(y_1,c_1), & c_1\ge\tau,\\
(y_2,c_2), & c_1<\tau\ \text{and}\ c_2\ge\tau,\\
(y_\varnothing,\min(c_1,c_2)), & c_1<\tau\ \text{and}\ c_2<\tau,
\end{cases}
\label{eq:inference-rule}
\end{equation}
where $(y_1,c_1)$ is the initial output and $(y_2,c_2)$ is generated
after re-attending to the most relevant visual region. Let
$A_1\in\mathbb{R}^{H\times W}$ be the cross-attention map aggregated over
answer tokens. The crop region is selected as
\begin{equation}
\mathcal{R}^{\star}=\operatorname*{argmin}_{\mathcal{R}}
|\mathcal{R}|\quad
\text{s.t.}\quad
\sum_{(u,v)\in\mathcal{R}} A_1(u,v)
\ge \kappa \sum_{u,v} A_1(u,v),
\label{eq:crop}
\end{equation}
with $\kappa=0.10$ by default. The cropped image
$I'=\operatorname{CropZoom}(I,\mathcal{R}^{\star})$ is then passed to the
same policy; no external verifier or second model is introduced. The
module adds at most one extra forward pass and only when the confidence
trigger fires. Since $c_{\mathrm{ans}}$ is trained to be calibrated, cases
with low confidence are concentrated among hard or ambiguous prompts, so
re-attention is rarely spent on already reliable answers.

\subsection{Training Pipeline Summary}
\label{sec:method:pipeline}

Training first performs schema SFT on $\sim$50K samples for one epoch to
initialise $o=(y,c_{\mathrm{ans}},\{(s_m,c_m)\})$ with pseudo labelled
confidences. It then runs GRPO with the calibration reward on $\sim$100K
prompts from standard VQA datasets and RLHF-V~\cite{yu2024rlhf}, using
$G{=}8$ rollouts,
$(\lambda_1,\lambda_2,\lambda_3,\lambda_4){=}(1.0,0.5,0.2,0.05)$ and
$\beta{=}0.5$. At inference, the parsed $c_{\mathrm{ans}}$ either returns
the answer directly or triggers one visual re-query when
$c_{\mathrm{ans}}<\tau$; we set $\tau{=}0.5$ by default.

\section{Experiments}
\label{sec:exp}

We organise the study around four questions, one per subsection:
\textbf{Q1}~Does \method reduce hallucinations across benchmarks and
backbones? (\S\ref{sec:exp:main}); \textbf{Q2}~Are its confidences
calibrated enough to support abstention? (\S\ref{sec:exp:calib});
\textbf{Q3}~How much does each component contribute?
(\S\ref{sec:exp:ablation}); \textbf{Q4}~Does it preserve general capability
at acceptable inference cost? (\S\ref{sec:exp:analysis}).

\paragraph{Backbones.}
We train and evaluate \method on two open source $\sim$8B instruct MLLMs:
\textbf{InternVL3-8B}~\cite{zhu2025internvl3} as the primary backbone and
\textbf{Qwen3-VL-8B-Instruct}~\cite{bai2025qwen3} for generalisation across
architectures, plus \textbf{LLaVA-OneVision-1.5-8B}~\cite{an2025llava} and
\textbf{MiniCPM-V~4.0} in a reduced transfer experiment
(\S\ref{sec:exp:analysis}).

\paragraph{Training data.}
The SFT initialisation uses 50K LLaVA-Instruct samples re-formatted with our
\texttt{<think>/<answer>/<confidence>} schema. Span confidences are pseudo
labelled from five stochastic decodes at $T{=}1.0$: a span scores high when
consistently supported across responses or matched by object/OCR
annotations, and low when it appears only in unsupported generations. Answer
targets come from answer agreement, or ground-truth correctness when
available. The RL stage uses $\sim$100K prompts: about 90K from VQAv2, GQA
and A-OKVQA plus 10K hallucination-targeted prompts from
M-HalDetect~\cite{gunjal2024detecting} and RLHF-V~\cite{yu2024rlhf}. To
avoid leakage we remove images overlapping any reported test set, and no
benchmark question or caption is used for pseudo labelling or rewards.

\paragraph{Hallucination benchmarks.}
We use four widely adopted suites. \textbf{POPE}~\cite{li2023evaluating}
reports F1 over yes/no object queries in three sampling regimes (random,
popular, adversarial). \textbf{HallusionBench}~\cite{guan2024hallusionbench}
stresses entangled language and visual illusions (aAcc, fAcc).
\textbf{AMBER}~\cite{wang2023amber} gives generative metrics, of which we
report AMBER-S F1 and CHAIR~\cite{rohrbach2018object}.
\textbf{MMHal-Bench}~\cite{sun2024aligning} reports an overall score (0--6)
and a hallucination rate, judged by GPT-4.

\paragraph{General capability and calibration metrics.}
To check that hallucination reduction does not cost overall ability, we also
evaluate MME, MMBench, MMStar and SEED-Bench. For every benchmark with
binary correctness we report \textbf{ECE} (15
bins)~\cite{guo2017calibration}, \textbf{Brier
score}~\cite{glenn1950verification}, the \textbf{AUROC} of $1{-}c$ as a
hallucination detector, and \textbf{Selective-Risk@90}, the error rate when
the model abstains on the $10\%$ least confident prompts.

\paragraph{Evaluation protocol.}
All calibration metrics use the parsed confidence $c_{\mathrm{ans}}$; if an
output omits the field we set $c_{\mathrm{ans}}{=}1.0$ and mark a format
error, penalising uncalibrated overconfidence rather than silently dropping
the sample. For open generation we follow the official AMBER and MMHal-Bench
scripts, using GPT-4 judging only for MMHal-Bench.

\paragraph{Baselines.}
We compare against the unmodified backbone, parameter tuning baselines
(SFT, HA-DPO, RLHF-V, mDPO, HALVA, POVID and vanilla GRPO without our
calibration reward), and inference baselines (VCD, OPERA, HALC and
Woodpecker~\cite{yin2024woodpecker}). All parameter tuning baselines are
re-trained on the same 100K prompt mix using their official
hyperparameters.

\paragraph{Implementation details.}
We apply LoRA~\cite{hu2022lora} of rank 64 to the language tower while
freezing the vision encoder; the connector is unfrozen during RL only.
Optimisation uses DeepSpeed ZeRO-3 + AdamW with learning rate
$\eta{=}1\!\times\!10^{-5}$ for one epoch of schema SFT and
$5\!\times\!10^{-6}$ for two epochs of GRPO. We set
$(\alpha,\gamma){=}(0.5,0.5)$ in Eq.~\eqref{eq:sft}, group size $G{=}8$,
KL coefficient $0.04$, clip $\epsilon{=}0.2$, and $K{=}15$ calibration
bins for the ECE bonus within each group. The reward weights are
$(\lambda_1,\lambda_2,\lambda_3,\lambda_4){=}(1.0,0.5,0.2,0.05)$,
with abstention parameters $(\tau_u,\rho_1,\rho_2){=}(0.4,0.5,0.3)$ and
length normalisation $L_{\max}{=}512$. Training is performed on
$8{\times}$ NVIDIA H800 (80\,GB) for $\sim$36 hours per backbone. The
re-attention threshold is fixed to $\tau{=}0.5$ unless otherwise specified;
the crop is the smallest axis aligned box covering the most attended region
that contains at least $10\%$ of the cross-attention mass. For
reproducibility we fix three random seeds and report the mean.

\subsection{Main Hallucination Results}
\label{sec:exp:main}

Table~\ref{tab:main} answers \textbf{Q1}. Across all four hallucination
benchmarks and both backbones, \method gives the best hallucination
suppression results among the parameter tuning methods in our comparison.
On InternVL3-8B, \method improves POPE F1 by $1.9$ points over RLHF-V and
by $1.5$ points over vanilla GRPO. It also raises the MMHal-Bench score
from $4.59$ to $4.85$ over GRPO while reducing AMBER CHAIR from $8.3$ to
$6.2$. The gains transfer cleanly to Qwen3-VL-8B, indicating that the
calibration reward is not tied to a particular backbone family. Decoding
baselines used at inference (VCD, OPERA) improve the base model without
additional training but plateau below \method, and---unlike \method---incur
their latency overhead at every deployment.

\begin{table}[t]
\caption{Main results across hallucination, calibration and general
capability benchmarks. Higher is better except for CHAIR and ECE. Bold =
best in column; underline = second best.}
\label{tab:main}
\centering
\scriptsize
\setlength{\tabcolsep}{3pt}
\begin{tabular}{lcccccccc}
\toprule
 & \multicolumn{2}{c}{POPE} & HalBench & \multicolumn{2}{c}{AMBER}
 & MMHal & MME & ECE \\
\cmidrule(lr){2-3}\cmidrule(lr){5-6}
Method & F1 $\uparrow$ & adv $\uparrow$ & aAcc $\uparrow$
 & F1 $\uparrow$ & CHAIR $\downarrow$
 & Score $\uparrow$ & P+C $\uparrow$ & $\downarrow$ \\
\midrule
\multicolumn{9}{l}{\emph{Backbone: InternVL3-8B}}\\
Base                       & 85.2 & 82.1 & 56.4 & 71.5 & 18.2 & 3.65 & 1554.2 & 0.175 \\
+SFT (instruct)            & 86.0 & 82.8 & 57.4 & 72.3 & 16.9 & 3.74 & 1548.5 & 0.183 \\
+VCD                       & 87.6 & 84.7 & 58.8 & 74.9 & 14.8 & 3.82 & 1554.2 & 0.175 \\
+OPERA                     & 87.1 & 85.0 & 60.2 & 74.1 & 13.7 & 3.96 & 1554.2 & 0.175 \\
+HALC                      & 88.4 & 85.5 & 61.4 & 75.0 & 12.4 & 4.15 & 1554.2 & 0.175 \\
+Woodpecker                & 88.1 & 86.3 & 60.9 & 75.7 & 12.9 & 4.08 & 1554.2 & 0.175 \\
+POVID                     & 88.9 & 86.5 & 62.0 & 76.2 & 11.6 & 4.21 & 1545.3 & 0.216 \\
+HALVA                     & 89.7 & 87.4 & 62.5 & 76.0 & 10.9 & 4.34 & 1542.8 & 0.221 \\
+mDPO                      & 89.3 & 87.9 & 63.4 & 77.1 & 10.6 & 4.27 & 1540.5 & 0.236 \\
+HA-DPO                    & 90.4 & 88.1 & 63.1 & 77.6 & 9.1 & 4.48 & 1538.2 & 0.242 \\
+RLHF-V                    & 90.5 & 88.7 & 64.6 & 77.9 & 8.8 & 4.52 & 1535.6 & 0.258 \\
+GRPO (w/o calib)          & \underline{90.9} & \underline{89.2} & \underline{65.3} & \underline{78.2} & \underline{8.3} & \underline{4.59} & 1550.4 & 0.272 \\
\textbf{+\method (Ours)}   & \textbf{92.4} & \textbf{91.1} & \textbf{68.7} & \textbf{80.5} & \textbf{6.2} & \textbf{4.85} & 1552.8 & \textbf{0.085} \\
\midrule
\multicolumn{9}{l}{\emph{Backbone: Qwen3-VL-8B-Instruct}}\\
Base                       & 86.5 & 83.8 & 58.5 & 73.2 & 16.4 & 3.82 & 1580.5 & 0.162 \\
+HA-DPO                    & 91.4 & 89.7 & 66.0 & 78.9 & 8.4 & 4.58 & 1565.2 & 0.238 \\
+RLHF-V                    & 91.6 & 90.4 & 65.7 & 79.3 & 7.9 & 4.66 & 1562.8 & 0.251 \\
+GRPO (w/o calib)          & \underline{92.1} & \underline{90.8} & \underline{66.8} & \underline{79.5} & \underline{7.2} & \underline{4.72} & 1578.4 & 0.265 \\
\textbf{+\method (Ours)}   & \textbf{93.6} & \textbf{92.5} & \textbf{70.4} & \textbf{82.1} & \textbf{5.1} & \textbf{5.05} & 1581.1 & \textbf{0.078} \\
\bottomrule
\end{tabular}
\end{table}

\subsection{Calibration and Selective Prediction}
\label{sec:exp:calib}

Table~\ref{tab:calib} answers \textbf{Q2} by isolating calibration
quality from raw accuracy. GRPO without calibration improves accuracy over
the backbone but, consistent with prior observations on RLHF-trained
policies~\cite{kadavath2022language,tian2023just}, degrades ECE: rewards
push the policy toward sharper, more overconfident outputs. \method
reverses this trend---ECE drops from $0.272$ to $0.085$ on POPE-adv and
from $0.288$ to $0.094$ on HallusionBench relative to GRPO without
calibration, while Brier and AUROC also improve. Selective-Risk@90 confirms
practical utility: at the same coverage, \method's residual error rate is
the lowest among all baselines, including DPO methods that do not expose
any usable confidence signal at all.

\begin{table}[t]
\caption{Calibration metrics on POPE-adv (left block) and
HallusionBench (right block). Lower is better for ECE/Brier/SR@90,
higher is better for AUROC.}
\label{tab:calib}
\centering
\scriptsize
\setlength{\tabcolsep}{2pt}
\renewcommand{\arraystretch}{0.95}
\begin{tabular}{@{}lcccc|cccc@{}}
\toprule
 & \multicolumn{4}{c|}{POPE-adv} & \multicolumn{4}{c}{HallusionBench} \\
Method & ECE $\downarrow$ & Brier $\downarrow$ & AUROC $\uparrow$ & SR@90 $\downarrow$
        & ECE $\downarrow$ & Brier $\downarrow$ & AUROC $\uparrow$ & SR@90 $\downarrow$\\
\midrule
Base                  & 0.175 & 0.142 & 0.645 & 12.8 & 0.192 & 0.165 & 0.621 & 18.5 \\
+HA-DPO               & 0.239 & 0.160 & 0.618 & 11.7 & 0.271 & 0.184 & 0.602 & 16.5 \\
+RLHF-V               & 0.258 & 0.164 & 0.605 & 11.1 & 0.268 & 0.190 & 0.589 & 16.9 \\
+GRPO (w/o calib)     & 0.272 & 0.171 & 0.588 & 10.8 & 0.288 & 0.195 & 0.565 & 15.7 \\
\textbf{+\method}     & \textbf{0.085} & \textbf{0.105} & \textbf{0.864} & \textbf{6.2}
                      & \textbf{0.094} & \textbf{0.122} & \textbf{0.845} & \textbf{9.4} \\
\bottomrule
\end{tabular}
\end{table}

\subsection{Ablation Study}
\label{sec:exp:ablation}

Table~\ref{tab:ablation} answers \textbf{Q3} on InternVL3-8B.
The most diagnostic finding is the gap between
rows ``Full \method'' and ``w/o $R_{\text{cal}}$'': removing the
calibration reward leaves the accuracy metrics largely unchanged but
increases ECE from $0.085$ to $0.268$. This isolates the contribution of
Eq.~\eqref{eq:calreward} from the standard accuracy signal: the
calibration term is responsible for the calibration gains rather than
for the accuracy gains. Dropping the abstention
reward $R_{\text{abs}}$ damages MMHal-Bench (where many prompts admit
``unknown'' as the right answer) but barely affects POPE. Dropping the
span confidence stream is harmless for closed form QA but hurts AMBER-F1,
indicating that local span confidences matter most for generative
hallucination metrics. Replacing the verbalised confidence with a learned
scalar head trades a small improvement in raw ECE for a noticeable accuracy
regression, which supports keeping confidence in plain text.

\begin{table}[t]
\caption{Component-level ablation on InternVL3-8B.}
\label{tab:ablation}
\centering
\scriptsize
\begin{tabular}{lccccc}
\toprule
Variant & POPE F1 $\uparrow$ & AMBER F1 $\uparrow$
        & MMHal $\uparrow$ & MME $\uparrow$ & ECE $\downarrow$ \\
\midrule
Full \method                                & 92.4 & 80.5 & 4.85 & 1552.8 & 0.085 \\
\;\;w/o span confidence                     & 92.2 & 77.4 & 4.73 & 1551.0 & 0.096 \\
\;\;w/o $R_{\text{cal}}$ (Eq.~\ref{eq:calreward}) & 91.8 & 80.1 & 4.82 & 1554.1 & 0.268 \\
\;\;\;\;w/o Brier sub-term only             & 92.1 & 80.0 & 4.79 & 1553.7 & 0.188 \\
\;\;\;\;w/o group-ECE bonus only            & 91.9 & 80.6 & 4.84 & 1552.9 & 0.139 \\
\;\;w/o $R_{\text{abs}}$                    & 92.3 & 80.2 & 4.35 & 1552.4 & 0.088 \\
\;\;w/o re-attention                        & 90.5 & 78.6 & 4.62 & 1552.8 & 0.085 \\
\;\;verbalised $\to$ confidence head        & 90.8 & 78.3 & 4.64 & 1541.9 & 0.081 \\
\;\;group size $G$: 8$\to$4                  & 91.6 & 79.1 & 4.76 & 1548.6 & 0.119 \\
\;\;group size $G$: 8$\to$16                 & 92.6 & 80.7 & 4.88 & 1553.5 & 0.082 \\
\bottomrule
\end{tabular}
\end{table}

\begin{figure}[t]
\centering
\begin{tikzpicture}
\begin{axis}[
    width=0.47\textwidth,
    height=4.0cm,
    x tick label style={font=\tiny},
    y tick label style={font=\tiny},
    xlabel={$\lambda_1$ (Calibration Reward Weight)},
    ylabel={POPE F1 (\%)},
    ymin=90, ymax=94,
    xtick={0, 0.25, 0.5, 1, 2},
    legend pos=south east,
    grid=both,
    grid style={dashed, gray!30},
    axis y line*=left,
    axis x line*=bottom,
    legend style={nodes={scale=0.7, transform shape}}
]
\addplot[color=blue, mark=*] coordinates {
    (0, 91.8) (0.25, 92.0) (0.5, 92.5) (1, 92.4) (2, 91.9)
};
\addlegendentry{POPE F1}
\end{axis}
\begin{axis}[
    width=0.47\textwidth,
    height=4.0cm,
    y tick label style={font=\tiny},
    axis y line*=right,
    axis x line=none,
    ylabel={ECE ($\downarrow$)},
    ymin=0.05, ymax=0.30,
    legend pos=north east,
    legend style={nodes={scale=0.7, transform shape}}
]
\addplot[color=red, mark=square*] coordinates {
    (0, 0.268) (0.25, 0.167) (0.5, 0.112) (1, 0.085) (2, 0.101)
};
\addlegendentry{ECE}
\end{axis}
\end{tikzpicture}\hfill
\begin{tikzpicture}
\begin{axis}[
    width=0.47\textwidth,
    height=4.0cm,
    tick label style={font=\tiny},
    xlabel={Confidence},
    ylabel={Empirical Accuracy},
    xmin=0, xmax=1,
    ymin=0, ymax=1,
    xtick={0, 0.2, 0.4, 0.6, 0.8, 1.0},
    ytick={0, 0.2, 0.4, 0.6, 0.8, 1.0},
    legend pos=north west,
    grid=both,
    grid style={dashed, gray!30},
    legend style={nodes={scale=0.75, transform shape}}
]
\addplot[color=gray, dashed, thick] coordinates {(0,0) (1,1)};
\addlegendentry{Perfect Calibration}

\addplot[color=halred, mark=triangle*, thick] coordinates {
    (0.1, 0.15) (0.3, 0.25) (0.5, 0.35) (0.7, 0.45) (0.9, 0.60)
};
\addlegendentry{GRPO (w/o calib)}

\addplot[color=halgreen, mark=*, thick] coordinates {
    (0.1, 0.12) (0.3, 0.28) (0.5, 0.48) (0.7, 0.72) (0.9, 0.88)
};
\addlegendentry{\method (Ours)}
\end{axis}
\end{tikzpicture}
\caption{\textbf{(a)} Sensitivity to the calibration weight $\lambda_1$: moderate
weighting best balances accuracy and calibration. \textbf{(b)} Reliability
diagram on POPE-adv; \method tracks the ideal diagonal more closely than
standard GRPO.}
\label{fig:diag}
\end{figure}
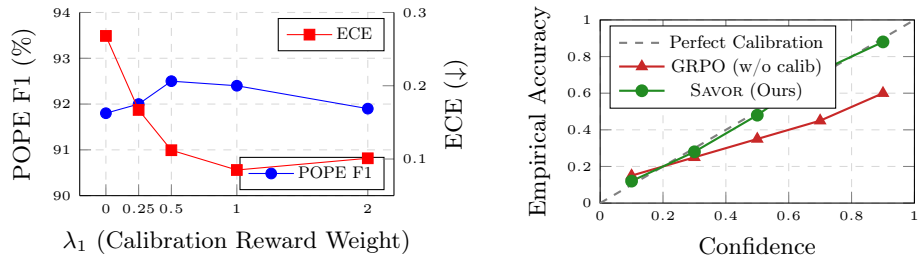

Fig.~\ref{fig:diag}(a) further examines reward weight sensitivity by
sweeping the calibration weight $\lambda_1\in\{0,0.25,0.5,1,2\}$ while
keeping the other reward weights fixed. Increasing the calibration weight
sharply reduces ECE at first, but an overly large weight starts to trade
away a small amount of POPE F1. The best trade-off is therefore obtained
around $\lambda_1{=}1.0$, with nearby settings showing similar behaviour.

\subsection{General Capability, Cost and Qualitative Analysis}
\label{sec:exp:analysis}

Table~\ref{tab:general} reports MME, MMBench, MMStar and SEED-Bench before
and after \method. Across four backbone families the changes are small with
no systematic downward trend: some metrics improve slightly, others drop by
less than the usual across-seed variance. This follows from the design, as
the RL stage keeps a small KL coefficient against the initialised policy,
preventing capability loss while still allowing large calibration shifts.

\begin{table}[t]
\caption{General capability benchmarks before / after \method.}
\label{tab:general}
\centering
\scriptsize
\begin{tabular}{lcccc}
\toprule
Backbone & MME P+C $\uparrow$ & MMBench $\uparrow$
         & MMStar $\uparrow$ & SEED $\uparrow$ \\
\midrule
InternVL3-8B (base)            & 1554.2 & 78.5 & 58.2 & 73.4 \\
\;\;+\method                   & 1552.8 & 78.7 & 58.0 & 73.6 \\
Qwen3-VL-8B (base)             & 1580.5 & 80.2 & 61.5 & 75.1 \\
\;\;+\method                   & 1581.1 & 80.0 & 61.8 & 74.8 \\
LLaVA-OneVision-1.5-8B (base)  & 1520.4 & 76.3 & 55.8 & 71.2 \\
\;\;+\method                   & 1519.2 & 76.6 & 55.7 & 71.4 \\
MiniCPM-V~4.0 (base)           & 1565.8 & 79.1 & 59.4 & 74.5 \\
\;\;+\method                   & 1564.7 & 78.9 & 59.6 & 74.6 \\
\bottomrule
\end{tabular}
\end{table}

Fig.~\ref{fig:diag}(b) plots predicted against empirical accuracy in 15
confidence bins. \method tracks the diagonal closely, while the base model
and uncalibrated GRPO are systematically overconfident in the $[0.7,1.0]$
range, the regime where hallucinations are most damaging.

\paragraph{Inference cost and analysis.}
With $\tau{=}0.5$, re-attention fires on $14.2\%$ of prompts and raises
latency by $16.5\%$, far below VCD ($\sim$$2{\times}$) and OPERA
($\sim$$1.7{\times}$), since the extra pass runs only on uncertain cases; it
is thus complementary to work that prunes multimodal inference
cost~\cite{wu2026vision}. Confidence also predicts correctness: on POPE-adv
and AMBER, Spearman $\rho$ rises from $0.24/0.21$ for prompted base
confidence to $0.67/0.63$. Remaining failures involve extra-image world
knowledge~\cite{yu2026believing}, compositional spatial reasoning that
crop-and-zoom cannot resolve~\cite{feng2026views}, or adversarial language
priors contradicting the image~\cite{qian2026penny}.

\section{Conclusion}
\label{sec:concl}

We presented \method, a calibration centred view of MLLM hallucination
mitigation. Rather than asking only which answer is preferred, we train the
model to estimate how reliable its own answer is and reward that estimate
inside a GRPO loop; the same confidence channel powers a re-attention step
at inference, closing a detect--correct loop within one policy. Across two
backbones and four benchmarks, \method cuts hallucinations and calibration
error together without sacrificing general capability. Next steps are richer
fusion and in-context alignment~\cite{meng2026adaptive,yang2026surface},
progressively trained smaller backbones~\cite{liu2026reasonact}, multi-agent
settings where confidence gates
delegation~\cite{meng2026group,kang2026mmlegal}, and stable calibration
under continual learning~\cite{feng2026forever}.

\bibliographystyle{splncs04}
\bibliography{savor}

\end{document}